\documentclass[10pt,twocolumn,letterpaper]{article}
\usepackage{cvpr}             
\usepackage{graphicx}
\usepackage{amsmath}
\usepackage{amssymb}
\usepackage{booktabs}
\usepackage{color}
\usepackage{colortbl}
\definecolor{Gray}{gray}{0.95}
\usepackage{adjustbox}
\usepackage{multirow}
\usepackage[pagebackref=true,breaklinks=true,colorlinks,bookmarks=false]{hyperref}
\usepackage{algorithm}
\usepackage{algpseudocode}
\usepackage{amsmath}
\usepackage{graphics}
\usepackage{epsfig}
\usepackage{soul}
\usepackage{algorithmicx,algorithm}
\usepackage{ulem}
\usepackage[capitalize]{cleveref}
\crefname{section}{Sec.}{Secs.}
\Crefname{section}{Section}{Sections}
\Crefname{table}{Table}{Tables}
\crefname{table}{Tab.}{Tabs.}

\newcommand{\paragrapha}[2][5pt]{\vspace{#1}\noindent\textbf{#2}}

\begin{document}
	
\title{Point-BERT: Pre-training 3D Point Cloud Transformers with \\ Masked Point Modeling}
	
	\author{
	Xumin Yu\thanks{Equal contribution. ~\textsuperscript{\dag}Corresponding author.} $  ^{,1}$,
	Lulu Tang$^{*,1,2}$, 
	Yongming Rao$^{*,1}$, 
	Tiejun Huang$^{2,3}$, 
	Jie Zhou$^{1}$, 
	Jiwen Lu$^{\dagger,1,2}$      \\
    $^1$Tsinghua University ~~	$^2$BAAI ~~  $^3$Peking University  \\
	}
	
	\maketitle
	\begin{abstract}
    We present \textit{Point-BERT}, a new paradigm for learning Transformers to generalize the concept of BERT\cite{bert} to 3D point cloud. Inspired by BERT, we devise a Masked Point Modeling (MPM) task to pre-train point cloud Transformers. Specifically,  we first divide a point cloud into several local point patches, and a point cloud \textit{Tokenizer} with a discrete Variational AutoEncoder (dVAE) is designed to generate discrete point tokens containing meaningful local information. Then, we randomly mask out some patches of input point clouds and feed them into the backbone Transformers. The pre-training objective is to recover the original point tokens at the masked locations under the supervision of point tokens obtained by the \textit{Tokenizer}. Extensive experiments demonstrate that the proposed  BERT-style pre-training strategy significantly improves the performance of standard point cloud Transformers. Equipped with our pre-training strategy, we show that a pure Transformer architecture attains 93.8\% accuracy on ModelNet40 and 83.1\% accuracy on the hardest setting of ScanObjectNN, surpassing carefully designed point cloud models with much fewer hand-made designs. We also demonstrate that the representations learned by Point-BERT transfer well to new tasks and domains, where our models largely advance the state-of-the-art of few-shot point cloud classification task. The code and pre-trained models are available at \url{https://github.com/lulutang0608/Point-BERT}.
	\end{abstract}
	
	\section{Introduction}	\label{sec:intro}
  Compared to conventional hand-crafted feature extraction methods, Convolutional Neural Networks (CNN)\cite{CNN} is dependent on much less prior knowledge. Transformers\cite{vaswani2017attention} have pushed this trend further as a step towards no inductive bias with minimal man-made assumptions, such as translation equivalence or locality in CNNs. Recently, the structural superiority and versatility of standard Transformers are proved in both language \cite{bert,brown2020language,radford2019language,joshi2020spanbert,liu2019roberta} and image tasks\cite{vit,zhu2020deformable,touvron2021training,xie2021self,chen2020generative,beit}, and the capability of diminishing the inductive biases is also justified by enabling more parameters, more data\cite{vit}, and longer training schedules. While Transformers produce astounding results in Natural Language Processing (NLP) and image processing, it is not well studied in the 3D community. Existing Transformer-based point cloud models\cite{zhao2021point,guo2021pct} bring in certain inevitable inductive biases from local feature aggregation\cite{zhao2021point} and neighbor embedding \cite{guo2021pct}, making them deviate from the mainstream of standard Transformers. To this end, we aim to apply standard Transformers on point cloud directly with minimal inductive bias, as a stepping stone to a neat and unified model for 3D representation learning. 
  
\begin{figure}[t]
\centering \includegraphics[width=\linewidth]{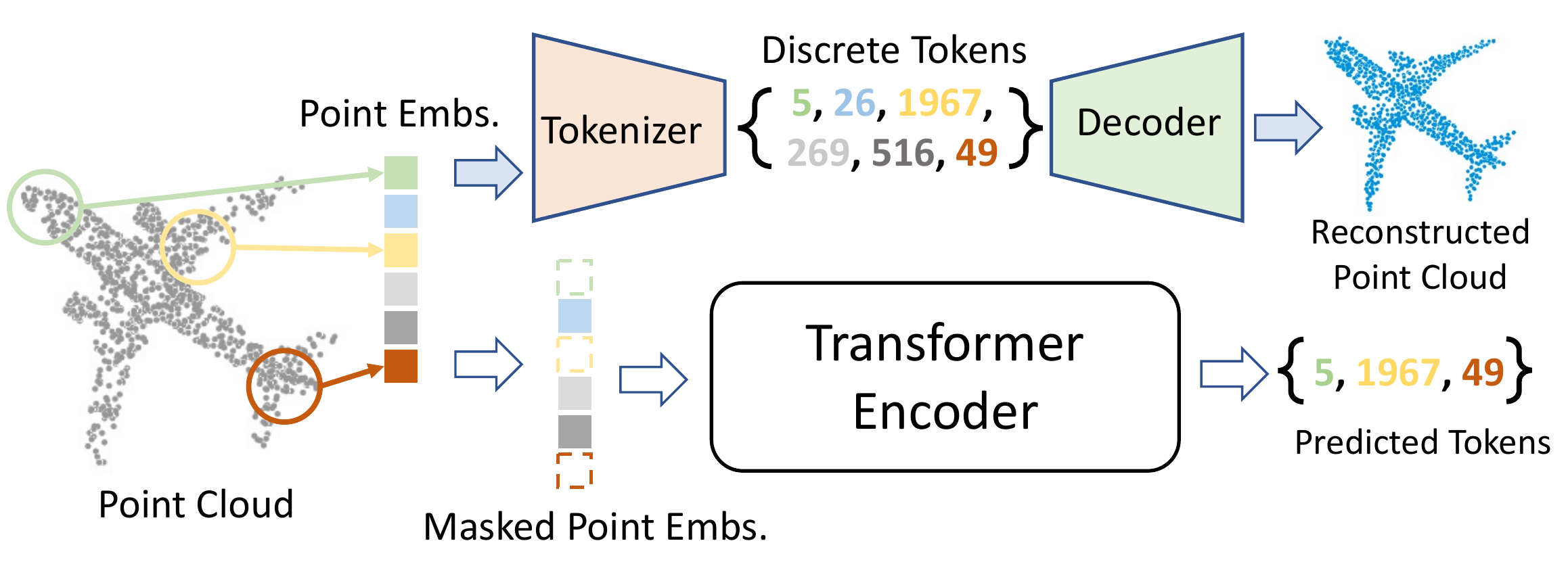}
\caption{\small \textbf{Illustration of our main idea.}  Point-BERT is designed for pre-training of standard point cloud Transformers. By training a dVAE via point cloud reconstruction, we can convert a point cloud into a sequence of discrete point tokens. Then we are able to pre-train the Transformers with a Mask Point Modeling (MPM) task by predicting the masked tokens. 
}
\vspace{-10pt}
\label{fig:insigt} 
\end{figure}

\begin{figure*}[t]
\centering
\includegraphics[width = 0.95\linewidth]{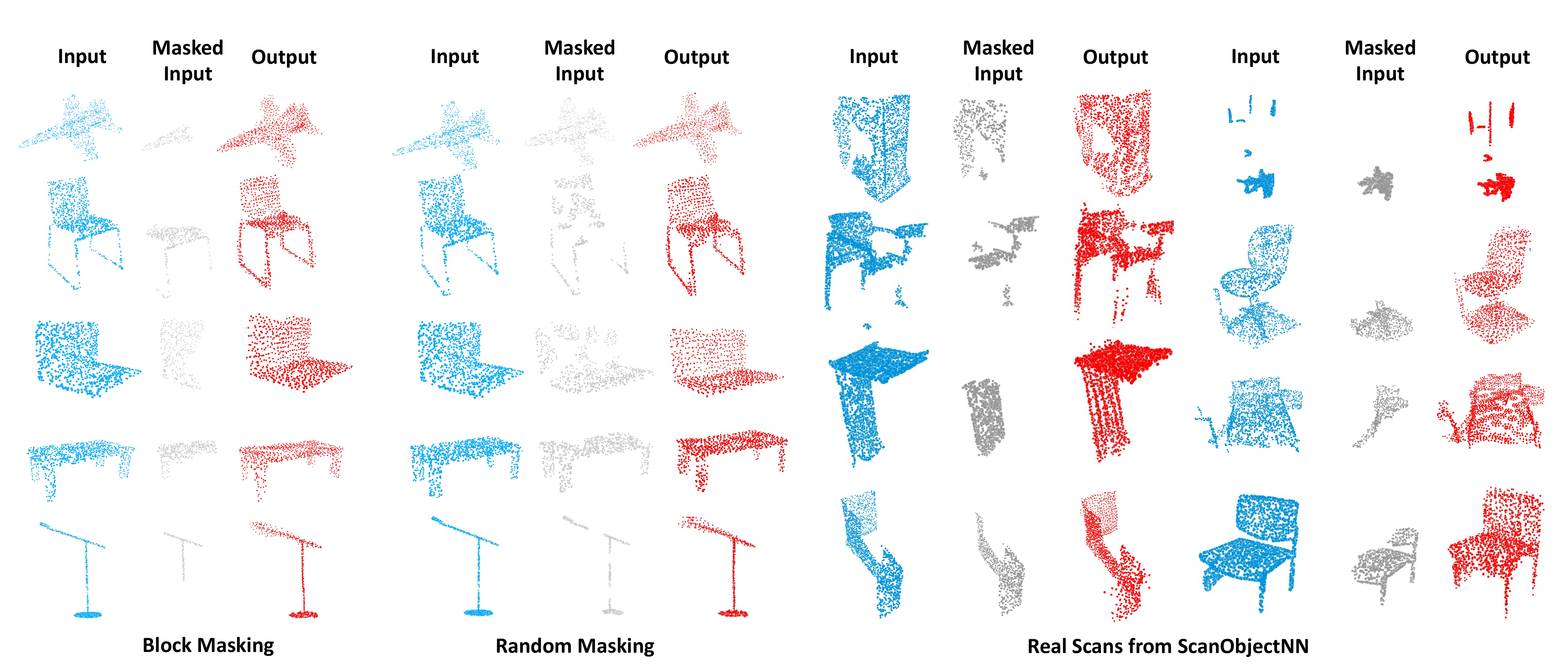}
\caption{\small \textbf{Masked point clouds reconstruction using our Point-BERT model trained on ShapeNet.} 
We show the reconstruction results of synthetic objects from ShapeNet test set with block masking and random masking in the first two groups respectively. Our model also generalize well to unseen real scans from ScanObjectNN (the last two groups). } 
\vspace{-7pt}
\label{fig:vis} 
\end{figure*}
		
Apparently, the straightforward adoption of Transformers does not achieve satisfactory performance on point cloud tasks (see Figure~\ref{fig:curve}). This discouraging result is partially attributed to the limited annotated 3D data since pure Transformers with no inductive bias need massive training data. For example, ViT\cite{vit} uses ImageNet\cite{CNN} (14M images) and JFT\cite{jft} (303M images) to train vision Transformers. In contrast, accurate annotated point clouds are relatively insufficient. Despite the 3D data acquisition is getting easy with the recent proliferation of modern scanning devices, labeling point clouds is still time-consuming, error-prone, and even infeasible in some extreme real-world scenarios. The difficulty motivates a flux of research into learning from unlabelled 3D data.  Self-supervised pre-training thereby becomes a viable technique to unleash the scalability and generalization of Transformers for 3D point cloud representation learning.

Among all the Transformer-based pre-training models, BERT\cite{bert} achieved state-of-the-art performance at its released time, setting a milestone in the NLP community. Inspired by BERT\cite{bert}, we seek to exploit the BERT-style pre-training for 3D point cloud understanding.  However, it is challenging to directly employ BERT on point clouds due to a lack of pre-existing vocabulary. In contrast, the language vocabulary has been well-defined (e.g., WordPiece in \cite{bert}) and off-the-shelf for model pre-training. In terms of point cloud Transformers, there is no pre-defined vocabulary for point clouds.  A naive idea is to treat every point as a `word' and mimic BERT\cite{bert} to predict the coordinates of masked points.  Such a point-wise regression task surges computational cost quadratically as the number of tokens increases. Moreover, a word in a sentence contains basic contextual semantic information, while a single point in a point cloud barely entails semantic meaning. 
	
Nevertheless, a local patch partitioned from a holistic point cloud contains plentiful geometric information and can be treated as a component unit.   
\textit{What if we build a vocabulary where different tokens represent different geometric patterns of the input units?} 
At this point, we can represent a point cloud as a sequence of such tokens. Now, we can favorably adopt BERT and its efficient implementations almost out of the box. We hypothesize that bridging this gap is a key to extending the successful Transformers and BERT to the 3D vision domain.

Driven by the above analysis, we present Point-BERT, a new scheme for learning point cloud Transformers. Two essential components are conceived: 1) Point Tokenization:  A  point cloud \textit{Tokenizer} is devised via a dVAE-based~\cite{rolfe2016discrete} point cloud reconstruction, where a point cloud can be converted into discrete point tokens according to the learned vocabulary. We expect that point tokens should imply local geometric patterns, and the learned vocabulary should cover diverse geometric patterns, such that a sequence of such tokens can represent any point cloud (even never seen before). 2) Masked Point Modeling: A `masked point modeling' (MPM) task is performed to pre-train Transformers, which masks a portion of input point cloud and learns to reconstruct the missing point tokens at the masked regions. We hope that our model enables reasoning the geometric relations among different patches of the point cloud, capturing meaningful geometric features for point cloud understanding.


Both two designs are implemented and justified in our experiments. 
We visualize the reconstruction results both on the synthetic (ShapeNet\cite{shapenet}) and real-world (ScanObjectNN\cite{uy2019revisiting}) datasets in Figure \ref{fig:vis}. We observe that Point-BERT correctly predicts the masked tokens and infers diverse, holistic reconstructions through our dVAE decoder. The results suggest that the proposed model has learned inherent and generic knowledge of 3D point clouds, i.e, geometric patterns or semantics. More significantly, our model is trained on  ShapeNet, the masked point predictions on ScanObjectNN reflect its superior performance on challenging scenarios with both unseen objects and domain gaps. 

 Our Point-BERT with a pure Transformer architecture and BERT-style pre-training technique achieves  93.8\% accuracy on ModelNet40 and 83.1\% accuracy on the complicated setting of ScanObjectNN, surpassing carefully designed point cloud models with much fewer human priors. We also show that the representations learned by Point-BERT transfer well to new tasks and domains, where our models largely advance the state-of-the-art of few-shot point cloud classification task. We hope a neat and unified Transformer architecture across images and point clouds could facilitate both domains since it enables joint modeling of 2D and 3D visual signals. 

\section{Related Work}
	

\paragrapha{Self-supervised Learning (SSL).} SSL is a type of unsupervised learning, where the supervision signals can be generated from the data itself\cite{hinton2020aaai}. The core idea of SSL is to define a pretext task, such as jigsaw puzzles \cite{jigsaw}, colorization\cite{colorization}, and optical-flow\cite{optical-flow} in images. More recently, several studies suggested using SSL techniques for point cloud understanding\cite{moco, infoce,pointcontrast,li2018so,Jigsaw3D,foldingnet,occo,rao2020global,eckart2021self,info3d,MD}. Example 3D pretext tasks includes orientation estimation\cite{Rotation3D},  deformation reconstruction\cite{achituve2021self}, geometric structural cues\cite{MortonNet} and spatial cues\cite{mersch2021self,sharma2020self}. Inspired by the jigsaw puzzles in images\cite{jigsaw}, \cite{Jigsaw3D} proposes to reconstruct point clouds from the randomly rearranged parts. A contrastive learning framework is proposed by DepthContrast \cite{DepthContrast} to learn representations from depth scans. More recently, OcCo\cite{occo} describes an encoder-decoder mechanism to reconstruct the occluded point clouds. Different from these studies, we attempt to explore a point cloud SSL model following the successful Transformers\cite{vaswani2017attention}.
	
\begin{figure*}[t]
\centering 
\includegraphics[width=\linewidth]{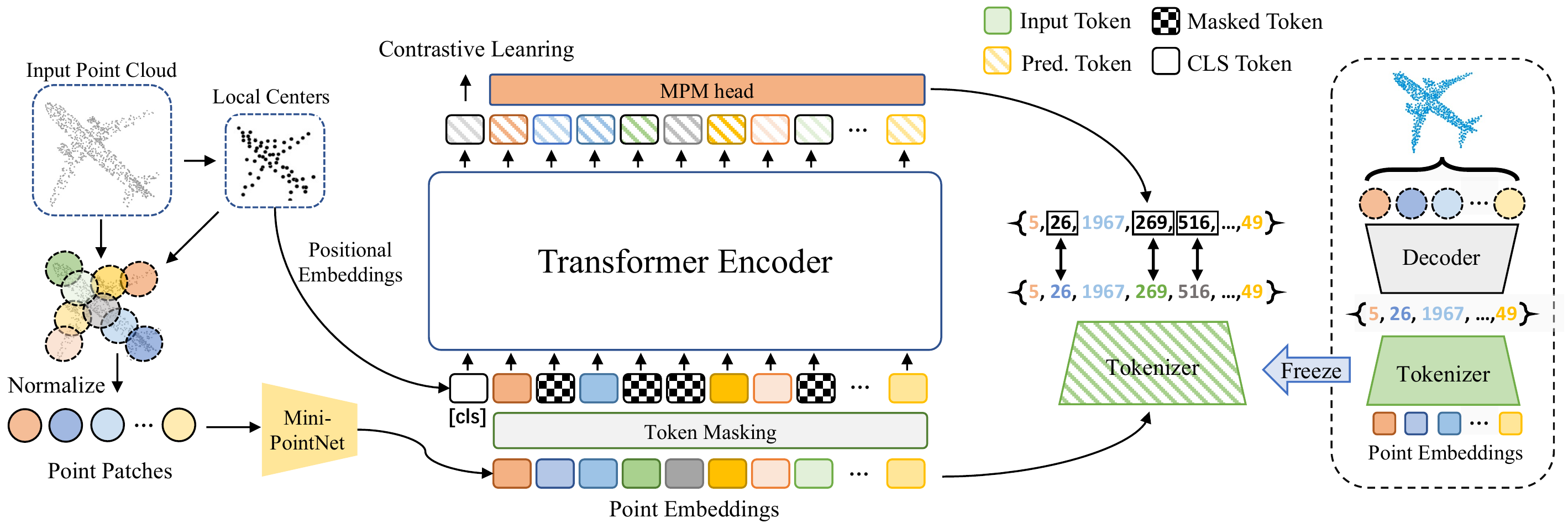}
\caption{ \small \textbf{The pipeline of Point-BERT.} We first partition the input point cloud into several point patches (sub-clouds).  A mini-PointNet\cite{pointnet} is then used to obtain a sequence of point embeddings. Before pre-training, a \textit{Tokenizer} is learned through dVAE-based point cloud reconstruction (as shown in the right part of the figure), where a point cloud can be converted into a sequence of discrete point tokens; During pre-training, we mask some portions of point embeddings and replace them with a mask token. The masked point embeddings are then fed into the Transformers. The model is trained to recover the original point tokens, under the supervision of point tokens obtained by the \textit{Tokenizer}. We also add an auxiliary  contrastive learning task to help the Transformers to capture high-level semantic knowledge.}
\label{fig:pipeline} \vspace{-10pt}
\end{figure*}
	
\paragrapha{Transformers.} Transformers\cite{vaswani2017attention} have become the dominant framework in NLP\cite{bert,brown2020language,radford2019language,joshi2020spanbert,liu2019roberta} due to its salient benefits, including massively parallel computing, long-distance characteristics, and minimal inductive bias. It has intrigued various vision tasks\cite{khan2021Transformers,han2020survey}, such as object classification\cite{vit,chen2020generative},  detection\cite{carion2020end,zhu2020deformable} and segmentation\cite{zheng2021rethinking,wang2021max}.
Nevertheless, its applications on point clouds remain limited. Some preliminary explorations have been implemented\cite{zhao2021point,guo2021pct,yu2021pointr}. For instance, \cite{zhao2021point} applies the vectorized self-attention mechanism to construct a point Transformer layer for 3D point cloud learning. \cite{guo2021pct} uses a more typical Transformer architecture with neighbor embedding to learn point clouds. Nevertheless, prior efforts for Transformer-based point cloud models more or less involve some inductive biases, making them out of the line with  standard Transformers.  In this work, we seek to continue the success of standard Transformers and extend it to point cloud learning with minimal inductive bias. 
	
\paragrapha{BERT-style Pre-training.} The main architecture of BERT\cite{bert} is built upon a  multi-layer Transformer encoder, which is first designed to pre-train bidirectional representations from the unlabeled text in a self-supervised scheme. The primary ingredient that helps BERT stand out and achieve impressive performance is the pretext of Masked Language Modeling (MLM), which first randomly masks and then recovers a sequence of input tokens.
	The MLM strategy has also inspired a lot of pre-training tasks\cite{joshi2020spanbert,liu2019roberta,conneau2019cross,trinh2019selfie,beit}. Take BEiT\cite{beit} for example, it first tokenizes the input image into discrete visual tokens. After that, it randomly masks some image patches and feeds the corrupted images into the Transformer backbone. The model is trained to recover the visual tokens of the masked patches.   More recently, MAE\cite{mae}  presents a masked autoencoder strategy for image representation learning. It first masks random patches of the input image and then encourages the model to reconstruct those missing pixels. Our work is greatly inspired by BEiT\cite{beit}, which encodes the image into discrete visual tokens so that a Transformer backbone can be directly applied to these visual tokens. However, it is more challenging to acquire tokens for point clouds due to the unstructured nature of point clouds, which subsequently hinders the straightforward use of BERT on point clouds. 

\section{Point-BERT}
\label{sec:app}
	
The overall objective of this work is to extend the BERT-style pre-training strategy to point cloud Transformers.	To achieve this goal, we first learn a  \textit{Tokenizer} to obtain discrete point tokens for each input point cloud.  Mimicking the `MLM' strategy in BERT\cite{bert}, we devise a `masked point modeling' (MPM) task to pre-train Transformers with the help of those discrete point tokens. The overall idea of our approach is illustrated in Figure~\ref{fig:pipeline}.

\subsection{Point Tokenization} \label{p:dvae}

\paragrapha{Point Embeddings.} 
A naive approach treats per point as one token. However, such a point-wise reconstruction task tends to unbearable computational cost due to the quadratic complexity of self-attention in Transformers. Inspired by the patch embedding strategy in Vision Transformers~\cite{vit}, we present a simple yet efficient implementation that groups each point cloud into several local patches (sub-clouds). Specifically, given an input point cloud $ p\in \mathbb{R}^{N\times 3}$, we first sample $g$ center points from the holistic point cloud $ p$ via farthest point sampling (FPS). The k-nearest neighbor (kNN) algorithm is then used to select the $n$ nearest neighbor points for each center point, grouping $g$ local patches (sub-clouds) $\left \{  p_{i} \right \}_{i=1}^{g}$.  We then make these local patches unbiased by subtracting their center coordinates, disentangling the structure patterns and spatial coordinates of the local patches. These unbiased sub-clouds can be treated as words in NLP or image patches in the vision domain. We further adopt a mini-PointNet\cite{pointnet} to project those sub-clouds into point embeddings. Following the practice of Transformers in NLP and 2D vision tasks, we represent a point cloud as a sequence of point embeddings $\left \{  f_{i} \right \}_{i=1}^{g}$, which can be received as inputs to standard Transformers.

\paragrapha{Point Tokenizer.}
Point \textit{Tokenizer}  takes point embeddings $\left \{  f_{i} \right \}_{i=1}^{g}$ as the inputs and converts them into discrete point tokens. Specifically, the \textit{Tokenizer} $\mathcal{Q}_{\phi}(z|f)$  maps point embeddings $\left \{  f_{i} \right \}_{i=1}^{g}$ into discrete point tokens $\mathbf{z}=[z_{1},z_{2},....,z_{g}] \in \mathcal{V} $~\footnote{Point tokens have two forms, discrete integer number and corresponding word embedding in $\mathcal{V}$, which are equivalent.}, where $\mathcal{V}$ is the learned vocabulary with total length of $N$. In this step, the sub-clouds  $\left \{  p_{i} \right \}_{i=1}^{g}$ can be tokenized into  point tokens $\left \{  z_{i} \right \}_{i=1}^{g}$, relating to  effective local geometric patterns. In our experiments, DGCNN\cite{wang2019dynamic} is employed as our \textit{Tokenizer} network.
	
\paragrapha{Point Cloud Reconstruction.}  The decoder  $\mathcal{P}_{\varphi }(p|z)$ of dVAE receives point tokens $\left \{  z_{i} \right \}_{i=1}^{g}$  as the inputs and learns to reconstruct the corresponding sub-clouds $\left \{  p_{i} \right \}_{i=1}^{g}$. Since the local geometry structure is too complex to be represented by the limited $N$ situations. We adopt a DGCNN\cite{wang2019dynamic} to build the relationship with neighboring point tokens, which can enhance the representation ability of discrete point tokens for diverse local structures. After that, a FoldingNet\cite{foldingnet} is used to reconstruct the sub-clouds.
	
The overall reconstruction objective can be written as $\mathbb{E}_{z\sim \mathcal{Q}_{\phi}(z|p)} \left [ \textup{ log  } \mathcal{P}_{\varphi  }(p|z) \right ]$, and the reconstruction procedure can be viewed as maximizing the evidence lower bound (ELB) of the log-likelihood $\mathcal{P}_{\theta }(p|\tilde{p})$\cite{ramesh2021zero}:
\small	
\begin{equation} \label{eq:elb}
\begin{split}	
	\sum_{(p_{i},\tilde{p}_{i})\in\mathcal{D}}\textup{log} \mathcal{P}_{\theta }(p_{i}|\tilde{p}_{i}) \geq   \sum_{(p_{i},\tilde{p}_{i})\in\mathcal{D}}(\mathbb{E}_{z_{i}\sim \mathcal{Q}_{\phi}(\mathbf{z}|p_{i})} \left [ \textup{ log  } \mathcal{P}_{\varphi  }(p_{i}|z_{i}) \right ]   \\
	- D_{\textup{KL}} [\mathcal{Q}_{\phi}(\mathbf{z}|p_{i}),\mathcal{P}_{\varphi }(\mathbf{z}|\tilde{p}_{i})  ] ),   \qquad \qquad \quad    (1)\nonumber%
\end{split}%
\end{equation}%
\normalsize
where $p$ denotes the original point cloud, $\tilde{p}$ denotes the reconstructed point cloud. Since the latent point tokens are discrete, we cannot apply the reparameterization gradient to train the dVAE. Following \cite{ramesh2021zero}, we use the Gumbel-softmax relaxation\cite{jang2016categorical} and a uniform prior during dVAE training. Details about dVAE architecture and its implementation can be found in the supplementary.

\subsection{Transformer Backbone}
We adopt the standard Transformers~\cite{vaswani2017attention} in our experiments, consisting of multi-headed self-attention layers and FFN blocks. For each input point cloud, we first divide  it into $g$ local patches with center points $\{c_i\}_{i=1}^g$.  Those local patches are then projected into point embeddings $\{f_i\}_{i=1}^g$  via a mini-PointNet~\cite{pointnet}, which consists of only MLP layers and the global \texttt{maxpool} operation. We further obtain the positional embeddings $\{pos_i\}$ of each patch by applying  an MLP  on its  center point $\{c_i\}$.  Formally, we define the input embeddings as $\{x_i\}_{i=1}^g$, which is the combination of point embeddings $\{f_i\}_{i=1}^g$ and positional embeddings $\{pos_i\}_{i=1}^g$. Then, we send the input embeddings  $\{x_i\}_{i=1}^g$ into the Transformer. Following~\cite{bert}, we append a class token $\mathbf{E}[\textup{s}]$ to the input sequences. Thus, the input sequence of Transformer can be expressed as $ H^{0}=\left \{ \mathbf{E}[\textup{s}],  x_{1},  x_{2}, \cdots ,  x_{g} \right \}$. There are $L$ layers of Transformer block, and the output of the last layer $ H^{L}= \left \{  h_{\textup{s}}^{L},  h_{1}^{L}, \cdots ,  h_{g}^{L}\right \}$ represents the global feature, along with the encoded representation of the input sub-clouds.

\subsection{Masked Point Modeling}

Motivated by BERT\cite{bert} and BEiT\cite{beit}, we extend the masked modeling strategy to point cloud learning and devise a masked point modeling (MPM) task for Point-BERT. 			

\paragrapha{Masked Sequence Generation.}  Different from the random masking used in  BERT~\cite{bert} and MAE\cite{mae}, we adopt a block-wise masking strategy like~\cite{beit}. Specifically, we first choose a center point $c_{i}$ along with its sub-cloud $p_{i}$, and then find its $m$ neighbor sub-clouds, forming a continuous local region. We mask out all local patches in this region to generate the masked point cloud.  In practice, we directly apply such a block-wise masking strategy like~\cite{beit} to the inputs of the Transformer.  Formally, we mark the masked positions as $\mathcal{M}\in \left \{ 1,\cdots ,g \right \}^{\lfloor rg  \rfloor}$, where $r$ is the mask ratio. Next, we replace all the masked point embeddings with a same learnable pre-defined mask embeddings $\mathbf{E}[\textup{M}]$ while keeping its positional embeddings unchanged. Finally, the corrupted input embeddings $
\textup{X}^{\mathcal{M}}=\left \{   x_{i}: i\notin \mathcal{M}   \right \}_{i=1}^{g}   \cup  \left \{ \mathbf{E}[\textup{M}] + {pos}_i :i\in  \mathcal{M} \right \}_{i=1}^{g} $ are fed into the Transformer encoder.

\paragrapha{Pretext Task Definition.}
The goal of our MPM task is to enable the model to infer the geometric structure of missing parts based on the remaining ones.  The pre-trained dVAE (see section~\ref{p:dvae})  encodes each local patch into discrete point tokens,  representing the geometric patterns. Thus, we can directly apply those informative tokens as our surrogate supervision signal to pre-train the Transformer. 
 
 \paragrapha{Point Patch Mixing. } Inspired by the CutMix\cite{cutmix,pointcutmix} technique, we additionally devise a neat mixed token prediction task as an auxiliary pretext task to increase the difficulty of pre-training in our Point-BERT, termed as `Point Patch Mixing'. Since the information of the absolute position of each sub-cloud has been excluded by normalization, we can create new virtual samples by simply mixing two groups of sub-clouds without any cumbersome alignment techniques between different patches, such as optimal transport~\cite{pointcutmix}. During pre-training, we also force the virtual sample to predict the corresponding tokens generated by the original sub-cloud to perform the MPM task. In our implementation, we generate the same number of virtual samples as the real ones to make the pre-training task more challenging, which is helpful to improve the training of Transformers with limited data as observed in~\cite{touvron2021training}.
	
\paragrapha{Optimization Objective. } The goal of MPM task is to recover the point tokens that are corresponding to the masked locations. The pre-training objective can be formalized as maximizing the log-likelihood of the correct point tokens $z_{i}$ given the masked input embeddings $\textup{X}^{\mathcal{M}}$:
\small\begin{equation}
	\textup{max}\sum_{\textup{X}\in D}^{}\mathbb{E}_{\mathcal{M}}\left [   \sum_{i\in \mathcal{M}  }^{} \textup{log} \mathcal{P}\left (  z_{i} | \textup{X}^{\mathcal{M}}\right ) \right ].  \quad    (2)
	\nonumber
\end{equation}\normalsize
MPM task encourages the model to predict the masked geometric structure of the point clouds.  Training the Transformer only with MPM task leads to an unsatisfactory understanding on high-level semantics of the point clouds, which is also pointed out by the recent work in 2D domain~\cite{ibot}. So we adopt the widely used contrastive learning method MoCo~\cite{moco} as a tool to help the Transformers to better learn high-level semantics. With our point patch mixing technique, the optimization of contrastive loss encourages the model to pay attention to the high-level semantics of point clouds by making features of the virtual samples as closely as possible to the corresponding features from the original samples. Let $q$ be the feature of a mixed sample that comes from two other samples, whose features are $k^{+}_1$ and $k^{+}_2$ ($\{k_i\}$ are extracted by the momentum feature encoder~\cite{moco}). Assuming the mixing ratio is $r$, the contrastive loss can be written as:
\small\begin{equation}
	\mathcal{L}_q = -r \textup{log} \frac{\textup{exp}(qk^{+}_1/ \tau)}{\sum_{i=0}^K \textup{exp}(qk_i/\tau)} - (1-r) \textup{log} \frac{\textup{exp}(qk^{+}_2/ \tau)}{\sum_{i=0}^K \textup{exp}(qk_i/\tau)}),    (3)
	\nonumber
\end{equation}\normalsize
where $\tau$ is the temperature and $K$ is the size of memory bank. Coupling MPM objective and contrastive loss enables our Point-BERT to simultaneously capture the local geometric structures and high-level semantic patterns, which are crucial in point cloud representation learning.

\section{Experiments}
In this section, we first introduce the setups of our pre-training scheme. Then we evaluate the proposed model with various downstream tasks, including object classification, part segmentation, few-shot learning and transfer learning. We also conduct an ablation study for our Point-BERT. 

\subsection{Pre-training Setups} 

\paragrapha{Data Setups.} ShapeNet~\cite{shapenet} is used as our pre-training dataset, which covers over 50,000 unique 3D models from 55 common object categories. We sample 1024 points from each 3D model and divide them into 64 point patches (sub-clouds). Each sub-cloud contains 32 points.  A lightweight PointNet\cite{pointnet} containing two-layer MLPs is adopted to project each sub-cloud into 64 point embeddings, which are used as input both for dVAE and Transformer.

\begin{table}[t]
\small
\caption{\textbf{Comparisons of Point-BERT with of state-of-the-art models on  ModelNet40. } We report the classification accuracy (\%) and the number of points in the input. [ST] and [T] represent the standard Transformers models and Transformer-based models with some special designs and more inductive biases, respectively.
}  \vspace{-5pt}
\centering
\label{tab:cls}
\setlength{\tabcolsep}{18pt}
\begin{tabular}{@{\hskip 5pt}>{\columncolor{white}[5pt][\tabcolsep]}lc>{\columncolor{white}[\tabcolsep][5pt]}l@{\hskip 5pt}}
\toprule
Method  &  $\#$point & Acc.\\
\midrule
PointNet\cite{pointnet} &  1k &89.2\\
PointNet++ \cite{pointnet2}&  1k &90.5\\
SO-Net\cite{li2018so} &  1k &92.5\\
PointCNN\cite{li2018pointcnn} &  1k &92.2\\
DGCNN\cite{wang2019dynamic} &  1k &92.9\\
DensePoint\cite{densepoint}&  1k &92.8\\
RSCNN\cite{rao2020global}&  1k &92.9\\
KPConv\cite{thomas2019kpconv} & $\sim$6.8k& 92.9\\
\midrule		
$\left [ \textup{T}\right ]$ PCT\cite{guo2021pct} &  1k & 93.2  \\
$\left [ \textup{T}\right ]$ PointTransformer\cite{zhao2021point} & -- & {93.7}  \\
$\left [ \textup{ST}\right ]$ NPCT\cite{guo2021pct} &  1k & 91.0 \\
\hline
$\left [ \textup{ST}\right ]$ Transformer &  1k & 91.4 \\
$\left [ \textup{ST}\right ]$ Transformer + OcCo~\cite{occo} &  1k & 92.1 \\
\rowcolor{Gray}  $\left [ \textup{ST}\right ]$ Point-BERT  & 1k & 93.2\\
\midrule
$\left [ \textup{ST}\right ]$ Transformer&  4k & 91.2  \\
$\left [ \textup{ST}\right ]$ Transformer + OcCo~\cite{occo} & 4k &92.2 \\
\rowcolor{Gray} $\left [ \textup{ST}\right ]$ Point-BERT & 4k & 93.4\\
\rowcolor{Gray} $\left [ \textup{ST}\right ]$ Point-BERT & 8k & \textbf{93.8}\\
\bottomrule
\end{tabular} 
\vspace{-15pt}
\end{table}
	
\paragrapha{dVAE Setups.} We use a four-layer DGCNN\cite{wang2019dynamic} to learn the inter-patch relationships, modeling the internal structures of input point clouds. During dVAE training, we set the vocabulary size $N$ to 8192. Our decoder is also a DGCNN architecture followed by a FoldingNet~\cite{foldingnet}. It is worth noting that the performance of dVAE is susceptible to hyper-parameters, which makes that the configurations of image-based dVAE~\cite{ramesh2021zero} cannot be directly used in our scenarios. The commonly used $\ell_1$-style Chamfer Distance loss is employed during the reconstruction procedure. Since the value of this $\ell_1$ loss is numerically small, the weight of KLD loss in Eq.\ref{eq:elb} must be smaller than that in the image tasks. We set the weight of KLD loss to 0 in the first 10,000 steps and gradually increased to 0.1 in the following 100,000 steps. The learning rate is set to 0.0005 with a cosine learning schedule with 60,000 steps warming up. We decay the temperature in Gumble-softmax function from 1 to 0.0625 in 100,000 steps following~\cite{ramesh2021zero}.  We train dVAE for a total of 150,000 steps with a batch size of 64.  

\paragrapha{MPM Setups.} In our experiments, we set the depth for the Transformer to 12, the feature dimension to 384, and the number of heads to 6. The stochastic depth~\cite{huang2016deep} with a 0.1 rate is applied in our transformer encoder. During MPM pre-training, we fix the weights of \textit{Tokenizer} learned by dVAE. 25\% $\sim$ 45\% input point embeddings are randomly masked out. The model is then trained to infer the expected point tokens at those masked locations. In terms of MoCo, we set the memory bank size to 16,384, temperature to 0.07, and weight momentum to 0.999. We employ an AdamW\cite{adamw} optimizer, using an initial learning rate of 0.0005 and a weight decay of 0.05.  The model is trained for 300 epochs with a batch size of 128. 
	
\subsection{Downstream Tasks}

\begin{table}[t] 
\caption{\small \textbf{Few-shot classification results on ModelNet40.} We report the average accuracy (\%) as well as the standard deviation over  10 independent experiments.}  \vspace{-5pt}
\newcolumntype{g}{>{\columncolor{Gray}}c}
\label{tab:few-shot}
\centering
\setlength{\tabcolsep}{4pt}{
\begin{adjustbox}{width=\linewidth} \small
\begin{tabular}{lcccc}
\toprule
\multirow{2}[0]{*}{} & \multicolumn{2}{c}{\textbf{5-way}} & \multicolumn{2}{c}{\textbf{10-way}} \\
\cmidrule(lr){2-3}\cmidrule(lr){4-5} 
& 10-shot & 20-shot & 10-shot & 20-shot \\
\midrule
DGCNN-rand\cite{occo} &31.6 $\pm$ 2.8 &  40.8 $\pm$ 4.6&  19.9 $\pm$  2.1& 16.9 $\pm$ 1.5\\
DGCNN-OcCo\cite{occo}&90.6 $\pm$ 2.8 & 92.5 $\pm$ 1.9 &82.9 $\pm$ 1.3 &86.5 $\pm$ 2.2 \\
\midrule
DGCNN-rand$^*$ &91.8 $\pm$ 3.7 & 93.4 $\pm$ 3.2 & 86.3 $\pm$ 6.2 &90.9 $\pm$ 5.1 \\
DGCNN-OcCo$^*$ & 91.9 $\pm$ 3.3& 93.9 $\pm$ 3.1  & 86.4 $\pm$ 5.4 & 91.3 $\pm$ 4.6\\
Transformer-rand & 87.8 $\pm$ 5.2& 93.3 $\pm$ 4.3 & 84.6 $\pm$ 5.5 & 89.4 $\pm$ 6.3\\
Transformer-OcCo& 94.0 $\pm$ 3.6& 95.9 $\pm$ 2.3 & 89.4 $\pm$ 5.1 & 92.4 $\pm$ 4.6 \\
\rowcolor{Gray}Point-BERT & \textbf{94.6 $\pm$ 3.1} & \textbf{96.3 $\pm$ 2.7} &  \textbf{91.0 $\pm$ 5.4} & \textbf{92.7 $\pm$ 5.1} \\
\bottomrule
\end{tabular}
\end{adjustbox}} 
\vspace{-15pt}
\end{table}

In this subsection, we report the experimental results on downstream tasks. Besides the widely used benchmarks, including classification and segmentation, we also study the model's capacity on few-shot learning and transfer learning.

\begin{table*}[t]
\caption{\small \textbf{Part segmentation results on the ShapeNetPart dataset}. We report the mean IoU across all part categories mIoU$_C$ (\%) and the mean IoU across all instance mIoU$_I$  (\%) , as well as the IoU (\%) for each categories. }   \vspace{-5pt}
\label{tab:ShapeNetPart}
\centering
\newcolumntype{g}{>{\columncolor{Gray}}c}
\setlength{\tabcolsep}{1.5mm}{
\begin{adjustbox}{width=\linewidth} \small
\begin{tabular}{l|c c|cccccccccccccccc}
\toprule
Methods& mIoU$_C$ & mIoU$_I$ & aero  & bag   & cap   & car   & chair & earphone & guitar & knife & lamp  & laptop & motor & mug   & pistol & rocket & skateboard  & table \\
\midrule
PointNet\cite{pointnet} & 80.39 & 83.7  & 83.4  & 78.7  & 82.5  & 74.9  & 89.6  & 73.0    & 91.5  & 85.9  & 80.8  & 95.3  & 65.2  & 93    & 81.2  & 57.9  & 72.8  & 80.6 \\
PointNet++\cite{pointnet2} & 81.85 & 85.1  & 82.4  & 79    & 87.7  & 77.3  & 90.8  & 71.8  & 91    & 85.9  & 83.7  & 95.3  & 71.6  & 94.1  & 81.3  & 58.7  & 76.4  & 82.6 \\
DGCNN\cite{wang2019dynamic} & 82.33 & 85.2  & 84    & 83.4  & 86.7  & 77.8  & 90.6  & 74.7  & 91.2  & 87.5  & 82.8  & 95.7 & 66.3  & 94.9  & 81.1  & 63.5  & 74.5  & 82.6 \\
\midrule
Transformer & 83.42 & 85.1  & 82.9  & 85.4  & 87.7  & 78.8  & 90.5  & 80.8  & 91.1  & 87.7 & 85.3  & 95.6  & 73.9  & 94.9 & 83.5  & 61.2  & 74.9  & 80.6 \\
Transformer-OcCo & 83.42 & 85.1  & 83.3  & 85.2  & 88.3 &  79.9 & 90.7  & 74.1  & 91.9  & 87.6  & 84.7  & 95.4  & 75.5 & 94.4  & 84.1  & 63.1  & 75.7  & 80.8 \\
\rowcolor{Gray} Point-BERT & \textbf{84.11} & \textbf{85.6}  & 84.3 & 84.8  & 88.0    & 79.8  & 91.0 & 81.7 & 91.6  & 87.9  & 85.2 & 95.6  & 75.6  & 94.7  & 84.3    & 63.4 & 76.3 & 81.5 \\
\bottomrule
\end{tabular}
\end{adjustbox}} \vspace{-7pt}
\end{table*}

\begin{table}[t]
\caption{\small \textbf{Classification results on the ScanObjectNN dataset.} We report the accuracy (\%) of three different settings. }  \vspace{-5pt}
\label{tab:ScanNN}
\centering
\newcolumntype{g}{>{\columncolor{Gray}}c}
\setlength{\tabcolsep}{1.5mm}{
\begin{adjustbox}{width=\linewidth} \small
\begin{tabular}{l | c c c}
\toprule
Methods & OBJ-BG & OBJ-ONLY & PB-T50-RS \\
\midrule 
PointNet\cite{pointnet}& 73.3  & 79.2  & 68.0 \\
SpiderCNN\cite{xu2018spidercnn} & 77.1  & 79.5  & 73.7 \\
PointNet++\cite{pointnet2} & 82.3  & 84.3  & 77.9 \\
PointCNN\cite{li2018pointcnn} & 86.1  & 85.5  & 78.5 \\
DGCNN\cite{wang2019dynamic} & 82.8  & 86.2  & 78.1 \\
BGA-DGCNN\cite{uy2019revisiting} & -- & -- & 79.7 \\
BGA-PN++\cite{uy2019revisiting} & -- & -- & 80.2 \\

\hline
Transformer & 79.86 & 80.55 & 77.24 \\
Transformer-OcCo & 84.85 & 85.54 & 78.79 \\
\rowcolor{Gray} Point-BERT & \textbf{87.43} & \textbf{88.12} & \textbf{83.07} \\
\bottomrule
\end{tabular}%
\end{adjustbox}} \vspace{-7pt}
\end{table}

\paragrapha{Object Classification.}
We conduct classification experiments on ModelNet40\cite{modelnet}, 
In the classification task, a two-layer MLP with a dropout of 0.5 is used as our classification head. We use AdamW with a weight decay of 0.05 and a learning rate of 0.0005 under a cosine schedule to optimize the model. The batch size is set to 32. 

The results are presented in Table \ref{tab:cls}. We denote our baseline model as  `Transformer', which is trained on ModelNet40  with random initialization.  Several Transformer-based models are illustrated, where [ST] represents a standard Transformer architecture, and [T] denotes the Transformer model with some special designs or inductive biases. Although we mainly focus on pre-training for standard Transformers in this work, our MPM pre-training strategy is also suitable for other Transformer-based point cloud models\cite{guo2021pct,zhao2021point}. Additionally, we compare with a recent pre-training strategy OcCo~\cite{occo}  as a strong baseline of our pre-training method. For fair comparisons, we follow the details illustrated in~\cite{occo} and use the Transfomer-based decoder PoinTr~\cite{yu2021pointr} to perform their pretext task.  Combining our Transformer encoder and PoinTr's decoder, we conduct the completion task on ShapeNet, following the idea of OcCo.  We term this model as `Transformer+OcCo'. 

We see pre-training Transformer with OcCo improves 0.7\%/1.0\% over the baseline using 1024/4096 inputs. In comparison, our Point-BERT brings 1.8\%/2.2\% gains over that of training from scratch. We also observe that adding more points will \textit{not} significantly improve the Transformer model without pre-training while Point-BERT models can be consistently improved by increasing the number of points. When we increase the density of inputs (4096),  our Point-BERT achieves significantly better performance (93.4\%) than that with the baseline (91.2\%) and OcCo (92.2\%).  Given more input points (8192), our method can be further boosted to 93.8\% accuracy on ModelNet40.

\paragrapha{Few-shot Learning.}
We follow previous work~\cite{sharma2020self} to evaluate our model under the few-shot learning setting. A typical setting is “$K$-way $N$-shot", where $K$ classes are first randomly selected, and then ($N$+20) objects are sampled for each class~\cite{sharma2020self}. The model is trained on $K\times N$ samples (support set), and evaluated on the remaining 20$K$ samples (query set). We compare Point-BERT with OcCo\cite{occo}, which achieves state-of-the-art performance on this task. In our experiments, we test the performance under ``5way 10shot", ``5way 20shot", ``10way 10shot" and ``10way 20shot". We conduct 10 independent experiments under each setting and report the average performance as well as the standard deviation over the 10 runs. We also reproduce DGCNN-rand and DGCNN-OcCo under the same condition for a fair comparison.

As shown in the Table~\ref{tab:few-shot}, Point-BERT achieves the best in the few-shot learning. It obtains an absolute improvement of 6.8\%, 3.0\%, 6.4\%, 3.3\% over the baseline and 0.6\%, 0.4\%, 1.6\%, 0.3\% over the OcCo-based method on the four settings. The strong results indicate that Point-BERT learns more generic knowledge that can be quickly transferred to new tasks with limited data.

\paragrapha{Part Segmentation.}
Object part segmentation is a challenging task aiming to predict a more fine-grained class label for every point.
We evaluate the effectiveness of Point-BERT on  ShapeNetPart~\cite{yi2016scalable}, which contains 16,881 models from 16 categories. Following PointNet~\cite{pointnet}, we sample 2048 points from each model and increase the group number $g$ from 64 to 128 in the segmentation tasks. We design a segmentation head to propagate the group features to each point hierarchically.
Specifically, features from $4^{th}$, $8^{th}$ and the last layer of Transformer are selected, denoted as $\{ H^4 = \{h_i^4\}_{i=1}^g , H^8 = \{h_i^8\}_{i=1}^g , H^{12} = \{h_i^{12}\}_{i=1}^g \}$.
 Then we downsample the origin point cloud to 512 and 256 points via FPS, phrased as $P^4 = \{p^4_i\}_{i=1}^{512}$ and $ P^8= \{p^8_i\}_{i=1}^{256}$. We follow PointNet++~\cite{pointnet2} to perform feature propagation between $H^4$ and $P^4$, $H^8$ and $P^8$. Here, we can obtain the upsampled feature map $\hat{H}^4$ and $\hat{H}^8$, which represent the features for the points in $P^4$ and $P^8$. Then, we can propagate the feature from $H^{12}$ to $\hat{H^4}$ and finally to every  point. 

Two types of mIoU are reported in Table~\ref{tab:ShapeNetPart}. It is clear that our Point-BERT outperforms PointNet, PointNet++, and DGCNN. Moreover, Point-BERT improves 0.69\% and 0.5\% mIoU over vanilla Transformers, while OcCo fails to improve baseline performance in part segmentation task.

\paragrapha{Transfer to Real-World Dataset.} We evaluate the generalization ability of the learned representation by pre-training the model on ShapeNet and fine-tuning it on ScanObjectNN~\cite{uy2019revisiting}, which contains 2902 point clouds from 15 categories. It is a more challenging dataset sampled from real-world scans containing background and occlusions. We follow previous works to conduct experiments on three main variants: OBJ-BG, OBJ-ONLY, and PB-T50-RS. The experimental results are reported in Table~\ref{tab:ScanNN}. 
As we can see, Point-BERT improves the vanilla Transformers by about 7.57\%, 7.57\%, and 5.83\% on three variants. 

Comparing the classification results  on ModelNet40 (Table~\ref{tab:cls}) and ScanObjectNN (Table~\ref{tab:few-shot}), we observe that DGCNN outperforms PointNet++ (+2.4\%) on the ModelNet40. While the superiority is degraded on the real-world dataset ScanObjectNN. As for Point-BERT, it achieves SOTA performance on both datasets, which strongly confirms the effectiveness of our method.

\begin{table}[t]
\small
\caption{\textbf{Ablation study.} We investigate the effects of different designs and report the classification accuracy (\%) after fine-tuning on ModelNet40. All models are trained with 1024 points. } \vspace{-5pt}
\centering
\label{tab:abl}
\newcolumntype{g}{>{\columncolor{Gray}}c}
\begin{adjustbox}{width=\linewidth}
\setlength{\tabcolsep}{5pt}
\begin{tabular}[\linewidth]{c | c c c |c }
\toprule
Pretext tasks  & MPM & Point Patch Mixing & Moco & Acc. \\
\midrule
Model A  & & & & 91.41 \\
Model B  &\checkmark & & & 92.58 $\uparrow$  \\
Model C    &\checkmark &\checkmark & & 92.91 $\uparrow$\\
\rowcolor{Gray} Model D     &\checkmark &\checkmark & \checkmark & 93.24 $\uparrow$ \\
\toprule
Augmentation  & mask type & mask ratio & replace  & Acc. \\
\midrule
\rowcolor{Gray} Model B & block mask & [0.25, 0.45]  & No & 92.58 \\
Model B & block mask & [0.25, 0.45]  & Yes & 91.81 $\downarrow$ \\
Model B & rand mask & [0.25, 0.45]  & No & 92.34  $\downarrow$ \\
Model B & block mask & [0.55, 0.85]  & No & 92.52 $\downarrow$\\
\midrule
\rowcolor{Gray} Model D & block mask & [0.25, 0.45] & No  & 93.16 \\
Model D & block mask & [0.25, 0.45] & Yes  & 92.58 $\downarrow$\\
Model D & rand mask & [0.25, 0.45] & No  & 92.91 $\downarrow$ \\
Model D & block mask & [0.55, 0.85] & No & 92.59 $\downarrow$ \\
\bottomrule
\end{tabular}
\end{adjustbox}\vspace{-7pt}
\end{table}

\subsection{Ablation Study}
\paragrapha{Pretext Task.} We denote model A as our baseline, which is the Transformer training from scratch. Model B presents pre-training Transformer with MPM pretext task. Model C is trained with more samples coming from `point patch mixing' technique. Model D (the proposed method) is trained under the setting of MPM, point patch mixing, and MoCo.  As can be seen in the upper part of Table~\ref{tab:abl}, Model B with MPM improves the performance about 1.17\% .  By adopting point patch mixing strategy, Model C gets an improvement of 0.33\%. With the help of MoCo\cite{moco}, Model D further brings an improvement of 0.33\%. 

\paragrapha{Masking Strategy.}  We visualize the point token prediction task in Figure~\ref{fig:vis}. Our Transformer encoder can reasonably infer the point tokens of the missing patches. In practice, we reconstruct the local patches through the decoder of dVAE, based on the point tokens predicted by the Transformer encoder. Two masking strategies are explored: block-wise masking (block-mask) and random masking (rand-mask). The masking strategy determines the difficulty of the pretext task, influencing reconstruction quality and representations. We further investigate the effects of different masking strategies and provide the results in Table~\ref{tab:abl}. We see that Model D with block-mask works better at the ratio of $ 25\% \sim 45 \% $.
Unlike images, which can be split into regular non-overlapping patches, sub-clouds partitioned from the original point cloud often involve overlaps. Thus, rand-mask makes the task easier than block-mask, and further degrades the reconstruction performance.  We also consider another type of augmentations: randomly replace some input embeddings with those from other samples.

\subsection{Visualization}

\begin{figure}[t]
\centering \includegraphics[width=\linewidth]{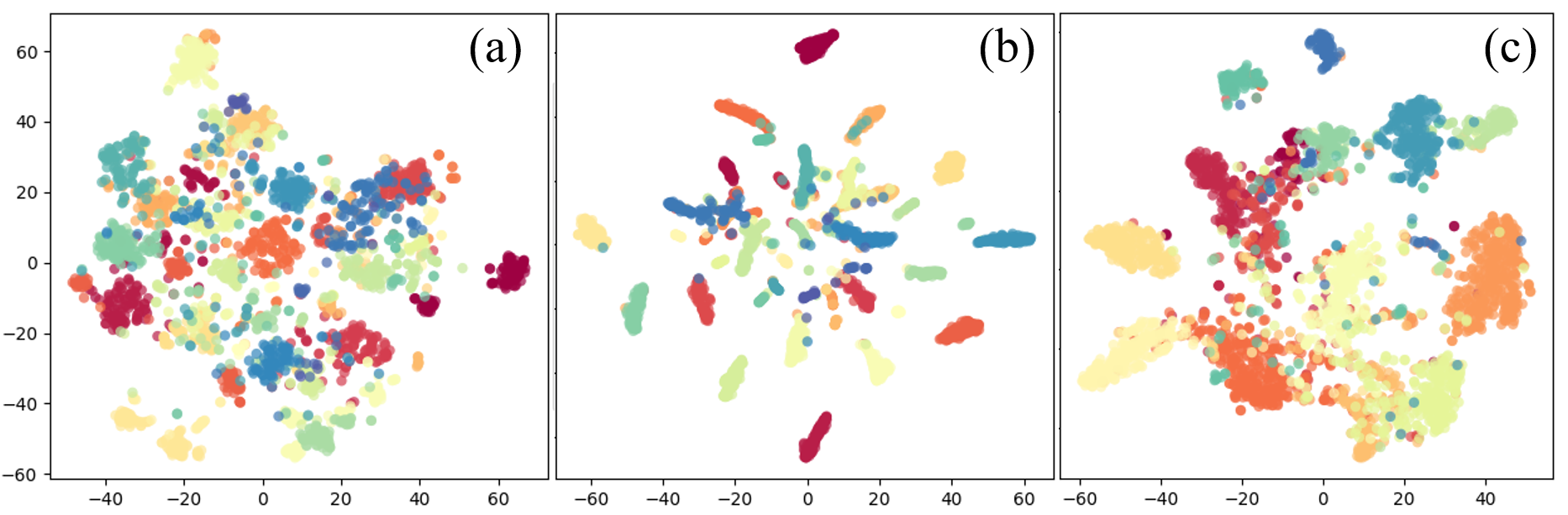}
\caption{\small \textbf{Visualization of feature distributions.} We show the t-SNE visualization of feature vectors learned by Point-BERT (a) after pre-training, (b)  after fine-tuning on ModelNet40, and (c) after fine-tuning on ScanObjectNN.}
\vspace{-10pt}
\label{fig:tsne}
\end{figure}

\begin{figure}[t]
\centering \includegraphics[width=\linewidth]{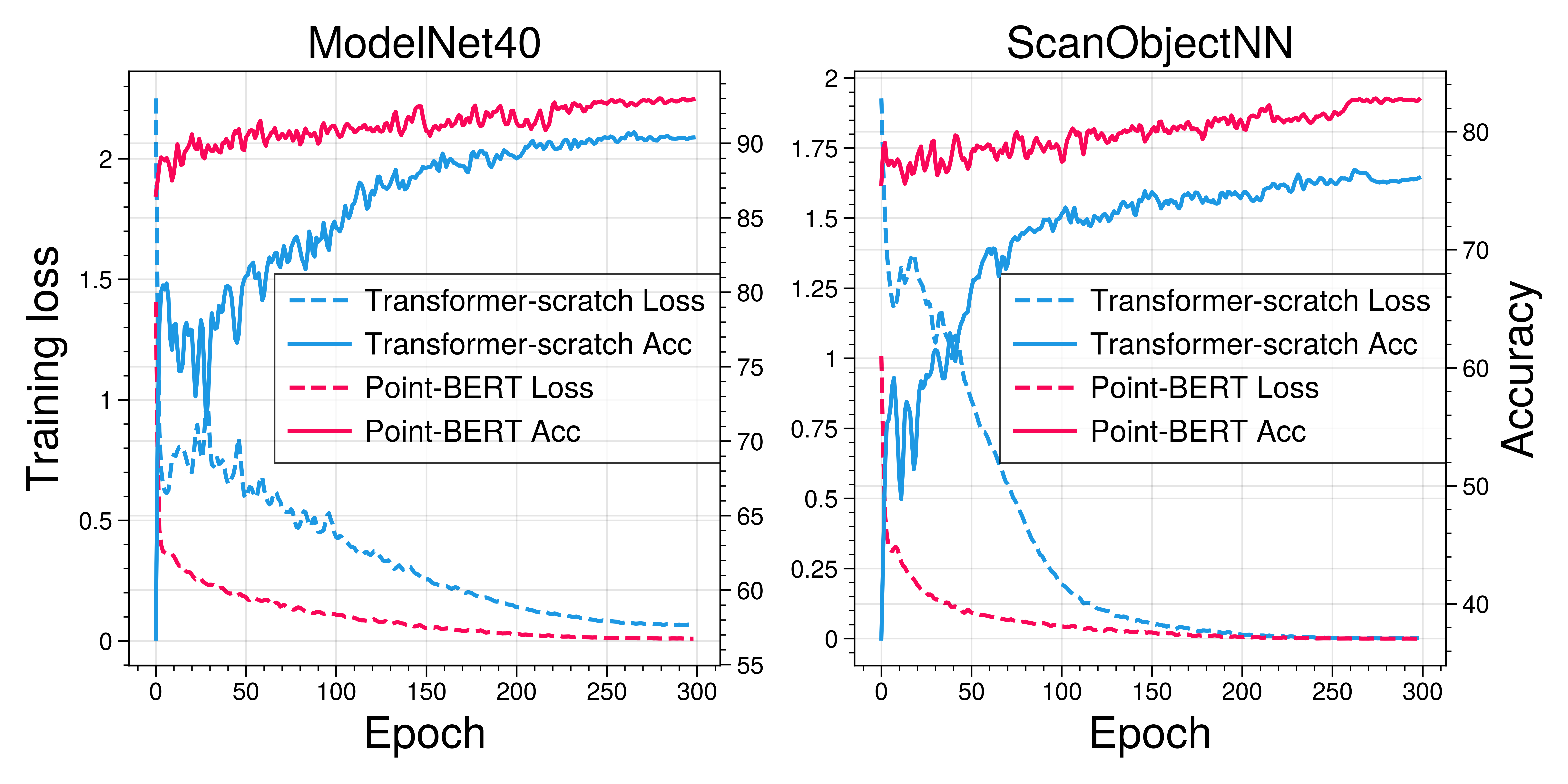}
\caption{ \textbf{Learning curve.} We compare the performance of Transformers training from scratch (blue) and pre-training with Point-BERT (red) in terms of training loss and validation accuracy on synthetic and real-world object classification datasets.
}
\vspace{-10pt}
\label{fig:curve}
\end{figure}

We visualize the learned features of two datasets via t-SNE~\cite{van2008visualizing} in Figure~\ref{fig:tsne}. In figure (a) and (b), the visualized features are from our Point-BERT (a) before fine-tuning and (b) after fine-tuning on ModelNet40. As can be seen, features from different categories can be well separated by our method even before fine-tuning.
We also visualize the feature maps on the PB-T50-RS of ScanObjectNN in (c). We can see that separate clusters are formed for each category, indicating the transferability of learned representation to real-world scenarios. It further verifies that Point-BERT helps the Transformer to learn generic knowledge for 3D objects. We also visualize the learning curves of our baseline Transformers and the proposed Point-BERT in Figure~\ref{fig:curve}. As can be seen, pre-training with our Point-BERT significantly improves the performance of baseline Transformers both in accuracy and speed on both synthetic and real-world datasets.

\section{Conclusion and Discussions}
We present a new paradigm for 3D point cloud Transformers through a BERT-style pre-training to learn both low-level structural information and high-level semantic feature. We observe a significant improvement for the Transformer on learning and generalization by comprehensive experiments on several 3D point cloud tasks. We show the potential of standard Transformers in 3D scenarios with appropriate pre-training strategy and look forward to further study on standard Transformers in the 3D domain. 

We do not foresee any negative ethical/societal impacts at this moment. Although the proposed method can effectively improve the performance of standard Transformers on point clouds, the entire `pre-training + fine-tuning' procedure is rather time-consuming, like other Transformers pre-training methods~\cite{bert, mae, beit}. Improving the efficiency of the training process will be an interesting future direction.

\subsection*{Acknowledgements}
This work was supported in part by the National Key Research and Development Program of China under Grant 2017YFA0700802, in part by the National Natural Science Foundation of China under Grant 62152603, Grant U1813218, in part by a grant from the Beijing Academy of Artificial Intelligence (BAAI), and in part by a grant from the Institute for Guo Qiang, Tsinghua University.

	{\small
    	\normalem
		\bibliographystyle{ieee_fullname}
		\bibliography{egbib}
	}

\newpage
\clearpage
\begin{appendix}
\section*{Appendix: Implementation Details}

\subsection*{A. Discrete VAE}
\noindent \textbf{Architecture:}
Our dVAE consists of a tokenizer and a decoder. Specifically, the tokenizer contains a 4-layer DGCNN~\cite{wang2019dynamic}, and the decoder involves a 4-layer DGCNN followed by a FoldingNet~\cite{foldingnet}. The detailed network architecture of our dVAE  is illustrated in Table~\ref{tab:arc}, where $C_{in}$ and $C_{out}$ are the dimension of input and output features, $C_{middle}$ is the dimension of the hidden layers. $N_{out}$ is the number of point patches in each layer, and  $\textup{K}$  is the number of neighbors in kNN operation. Additionally, $\texttt{FoldingLayer}$ concatenates a 2D grids to the inputs and finally generates 3D point clouds. 

\begin{table}[ht]
\small
\caption{\textbf{Detailed architecture of our models.} $C_{in}$/$C_{out}$  represents the dimension of input/output features, and $N_{out}$ is the number of points in the query point cloud. $\textup{K}$ is the number of neighbors in kNN operation. $C_{middle}$ is the dimension of the hidden layers for MLPs.} 
\centering
\label{tab:arc}
\setlength{\tabcolsep}{1.5mm}{
\begin{adjustbox}{width=0.45\textwidth}
\begin{tabular}{l | c |  c |c | c |c |c }
\toprule
Module  & Block & $C_{in}$  &  $C_{out}$ & K & $N_{out}$ & $C_{middle}$  \\
 \midrule
                 &  Linear & 256 & 128 &  &  &    \\
                 &  DGCNN & 128  & 256 & 4 & 64 &    \\ 
dVAE Tokenizer   &  DGCNN & 256  & 512 & 4 & 64 &    \\ 
                 &  DGCNN & 512  & 512 & 4 & 64 &   \\ 		
                 &  DGCNN & 512  & 1024 & 4 & 64 &   \\ 
                 &  Linear & 2304 & 8192 &  &  &    \\ 
\midrule[1pt]		
                &  Linear & 256 & 128 &  &  &    \\
                &  DGCNN & 128  & 256 & 4 & 64 &    \\ 
                &  DGCNN & 256  & 512 & 4 & 64 &    \\ 
  dVAE Decoder    &  DGCNN & 512  & 512 & 4 & 64 &    \\ 		
    &  DGCNN & 512  & 1024 & 4 & 64 &    \\ 
                &  Linear & 2304 & 256 &  &  &    \\
                &  MLP   & 256 & 48 &  &  & 1024   \\
                &  FoldingLayer  & 256 & 3 &  &  & 1024 \\
\midrule[1pt]
Classification Head   &  MLP  & 768 & $N_{cls}$ &  &  & 256 \\
\midrule[1pt] 		
              &  MLP  & 387 & 384 &  &  & 384$\times$4\\
         	  &  DGCNN & 384  & 512 & 4 & 128 &    \\ 		
              &  DGCNN & 512  & 384 & 4 & 128 &    \\ 
                &  DGCNN & 384  & 512 & 4 & 256 &    \\ 		
Segmentation Head   &  DGCNN & 512  & 384 & 4 & 256 &    \\  	
                &  DGCNN & 384 & 512 & 4 & 512 &    \\
                &  DGCNN & 512 & 384 & 4 & 512 &    \\ 
                &  DGCNN & 384 & 512 & 4 & 2048 &    \\ 
              &  DGCNN & 512 & 384 & 4 & 2048 &   \\ 
\bottomrule
\end{tabular}
\end{adjustbox}}
\end{table}

\paragrapha{Optimization:}
During the training phase, we consider reconstruction loss and distribution loss simultaneously. For reconstruction, we follow PoinTr~\cite{yu2021pointr} to supervise both coarse-grained prediction and fine-grained prediction with the ground-truth point cloud. The $\ell_1$-form Chamfer Distance is adopted, which is calculated as:
\small
\begin{eqnarray} \centering
d_{CD}^{\ell_1}(\mathcal{P},\mathcal{G}) = \frac{1}{|\mathcal{P}|}\sum_{p\in \mathcal{P}} \min_{g\in \mathcal{G}} \|p-g\| + \frac{1}{|\mathcal{G}|}\sum_{g\in \mathcal{G}} \min_{p\in \mathcal{P}} \|g-p\|,  
\end{eqnarray}
\normalsize
where $\mathcal{P}$ represents the prediction point set and $\mathcal{G}$ represents the ground-truth point set. 
Except for the reconstruction loss, we follow~\cite{ramesh2021zero} to optimize the KL-divergence $\mathcal{L}_{KL}$ between the predicted tokens' distribution and a uniform prior. The final objective function is 
\begin{small}
\begin{eqnarray} \centering
\mathcal{L}_{\text{dVAE}} = d_{CD}^{\ell_1}(\mathcal{P}_{fine},\mathcal{G}) + d_{CD}^{\ell_1}(\mathcal{P}_{coarse},\mathcal{G}) + \alpha \mathcal{L}_{KL}. 
\end{eqnarray}
\end{small}

\paragrapha{Experiment Setting:}
We report the default setting for dVAE training in Table~\ref{tab:supp_dvae_setting}.

\begin{table}[t]
\caption{\small \textbf{Experiment setting for training the dVAE.}} 
\setlength{\tabcolsep}{20pt}
\begin{tabular}{@{\hskip 5pt}>{\columncolor{white}[5pt][\tabcolsep]}l|c @{\hskip 5pt}>{\columncolor{white}[\tabcolsep][5pt]}cl}
config & value\\
\midrule[1.5pt]
optimizer & AdamW~\cite{loshchilov2018fixing}\\
learning rate & 5e-4\\
weight decay & 5e-4\\
learning rate schedule & cosine~\cite{loshchilov2016sgdr}\\
warmingup epochs & 10\\
augmentation & RandSampling\\
batch size & 64\\
number of points & 1024\\
number of patches & 64\\
patch size & 32\\
training epochs & 300\\
dataset & ShapeNet~\cite{shapenet}\\

\end{tabular} 
\label{tab:supp_dvae_setting}
\centering
\end{table}

\paragrapha{Hyper-parameters of dVAE:} We set the size of the learnable vocabulary to 8192, and each `word' in it is a 256-dim vector. The most important and sensitive hyper-parameters of dVAE are $\alpha$ for $\mathcal{L}_{KL}$  and the temperature $\tau$ for Gumbel-softmax. We set $\alpha$ to 0 in the first 18 epochs (about 10,000 steps) and gradually increase to 0.1 in the following 180 epochs (about 100,000 steps) using a cosine schedule. As for  $\tau$, we follow~\cite{ramesh2021zero} to decay it from 1 to 0.0625 using a cosine schedule in the first 180 epochs (about 100,000 steps). 
\subsection*{B. Point-BERT}
\noindent \textbf{Architecture: } We follow the standard Transformer~\cite{vit} architecture in our experiments. It contains a stack of Transformer blocks~\cite{vaswani2017attention}, and each block consists of a multi-head self-attention layer and a FeedForward Network (FFN). In these two layers, LayerNorm (LN) is adopted.

\paragrapha{Multi-head Attention: } Multi-head attention mechanism enables the network to jointly consider information from different representation subspaces~\cite{vaswani2017attention}. Specifically, given the input values $V$, keys $K$ and queries $Q$, the multi-head attention is computed by:
\begin{small}
\begin{equation}
    \begin{split}
        \text{MultiHead}(Q, K, V) = W^o \text{Concat}(\text{head}_1, ..., \text{head}_h),
    \end{split}
\end{equation}
\end{small}

\noindent where $W^o$ is the weights of the last linear layer. The feature of each head can be obtained by:
\begin{equation}
    \begin{split}
        \text{head}_i = \text{softmax}(\frac{QW_i^Q (KW_i^K)^T}{\sqrt{d_k}}) VW_i^V , 
    \end{split}
\end{equation}

\noindent where $W_i^Q$, $W_i^K$ and $W_i^V$ are the linear layers that project the inputs to different subspaces and $d_k$ is the dimension of the input features.  

\paragrapha{Feed-forward network (FFN): } Following~\cite{vaswani2017attention}, two linear layers with ReLU activations and dropout are adopted as the feed-forward network. 

\begin{table}[t]

\setlength{\tabcolsep}{20pt}
\begin{tabular}{@{\hskip 5pt}>{\columncolor{white}[5pt][\tabcolsep]}l|c @{\hskip 5pt}>{\columncolor{white}[\tabcolsep][5pt]}cl}
config & value\\
\midrule[1.5pt]
optimizer & AdamW\\
learning rate & 5e-4\\
weight decay & 5e-2\\
learning rate schedule & cosine\\
warmingup epochs & 3\\
augmentation & ScaleAndTranslate\\
batch size & 128\\
number of points & 1024\\
number of patches & 64\\
patch size & 32\\
mask ratio & [0.25, 0.45]\\
mask type & rand mask\\
training epochs & 300\\
dataset & ShapeNet\\

\end{tabular} 
\small
\caption{\textbf{Experiment setting for Point-BERT pre-training}}
\label{tab:pretrain}
\centering
\end{table}

\begin{table}[t]
\setlength{\tabcolsep}{20pt}
\begin{tabular}{@{\hskip 5pt}>{\columncolor{white}[5pt][\tabcolsep]}l|c >{\columncolor{white}[\tabcolsep][5pt]}c@{\hskip 5pt}}
config & value\\
\midrule[1.5pt]
optimizer & AdamW\\
learning rate & 5e-4\\
weight decay & 5e-2\\
learning rate schedule & cosine\\
warmingup epochs & 10\\
augmentation & ScaleAndTranslate\\
batch size & 32($C$),16($S$)\\
number of points & 1024($C$),2048 ($S$)\\
number of patches & 64($C$),128($S$)\\
patch size & 32\\
training epochs & 300\\
\end{tabular} 
\small
\caption{\textbf{Experiment setting for end-to-end finetuning.} $S$ represents segmentation task, $C$ represents classification task. } 
\label{tab:finetuning}
\centering
\end{table}

\paragrapha{Point-BERT pre-training:}
We report the default setting for our experiments in Point-BERT pretraining in Table~\ref{tab:pretrain}. The pre-training is conducted on ShapeNet.

\paragrapha{End-to-end finetuning:}
We finetune our Point-BERT model follow the common practice of supervised models strictly. The default setting for end-to-end finetuning is in Table~\ref{tab:finetuning}.

\paragrapha{Hyper-parameters of Transformers:}  
We set the number of blocks in the Transformer to 12. The number of heads in each multi-head self-attention layer is set to 6. The feature dimension of the transformer layer is set to 384. We follow~\cite{touvron2021training} to adopt the stochastic depth strategy with a drop rate of 0.1.

\paragrapha{Classification Head:}
A two-layer MLP with dropout is applied as our classification head. In classification tasks, we first take the output feature of [CLS] token out, and max-pool the rest of nodes' features. These two features are then combined together and sent into the classification head. The detailed architecture of the classification head is shown in Table~\ref{tab:arc}, where $N_{cls}$ is the number of classes for a certain dataset.

\paragrapha{Segmentation Head:}
There are no downsampling layers in the standard Transformers, making it challenging to perform dense prediction based on a single-resolution feature map. We adopt an upsampling-propagation strategy to solve this problem, consisting of two steps: 1) Geometry-based feature upsampling and 2) Hierarchical feature propagation. 

\begin{figure}[t]
\centering
\includegraphics[width = \linewidth]{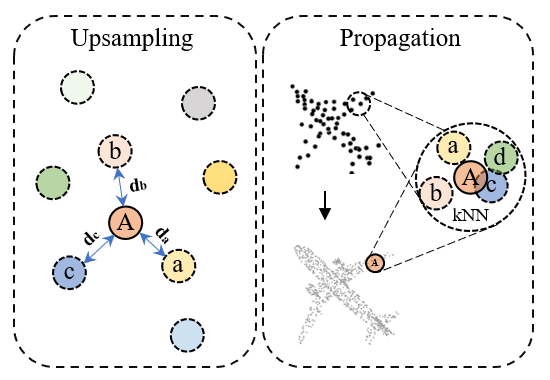}
\caption{\small \textbf{Two main operations of our segmentation head:} 1) Upsampling: upsample the feature map for the sparse point cloud to the dense point cloud. 2) Propagation: propagate the feature hierarchically from deep layers to shallow layers for dense prediction.}
\label{fig:supp_seg} 
\end{figure}

We extract features from different layers of the Transformer, where features from shallow layers tend to capture low-level information,  while features from deeper layers involve more high-level information. To upsample the feature maps to different resolutions, we first apply FPS to the origin point cloud and obtain point clouds at various resolutions. Then we upsample the feature maps from different layers to different resolutions accordingly. As shown in the left part of Figure~\ref{fig:supp_seg}, `A' is a point from the dense point cloud, and `a',`b',`c' are its nearest points in the sparser point cloud,  with distance of $d_a$, $d_b$ and $d_c$ respectively. We obtain the point feature of `A' based on the weighted addition of those features, which can be written as:

\begin{equation}
    \begin{split}
        \mathcal{F}_A = \text{MLP} (\text{Concat}(\frac{\sum_{i \in [a,b,c]} \frac{1}{d_i} \mathcal{F}_{i}}{\sum_{i \in [a,b,c]} \frac{1}{d_i}}, p_A)), 
    \end{split}
\end{equation}
\noindent where $p_A$ represents the coordinates of point `A'.

After obtaining the feature maps at different resolutions, we perform feature propagation from coarse-grained feature maps to fine-grained feature maps. As shown in the right part of Figure~\ref{fig:supp_seg}, for a point `A' in the dense point cloud, we find its $k$ nearest points in the sparser point cloud. Then a lightweight DGCNN~\cite{wang2019dynamic} is used to update the feature of `A'. We hierarchically update the feature with the resolution increases and finally obtain the dense feature map, which can be used for segmentation tasks. The detailed architecture for the segmentation head is shown in Table~\ref{tab:arc}. 

\end{appendix}
	
\end{document}